\documentclass[
]{ceurart}

\usepackage{linguex}

\usepackage[frozencache=true,cachedir=minted-cache]{minted} 

\setminted{breaklines=true}

\begin{document}

\copyrightyear{2022}
\copyrightclause{Copyright for this paper by its authors.
  Use permitted under Creative Commons License Attribution 4.0
  International (CC BY 4.0).}

\conference{The International Conference on Agglutinative Language Technologies as a challenge of Natural Language Processing, ALTNLP'22, June 7-8, Koper, Slovenia}

\title{A description of Turkish Discourse Bank 1.2 and an examination of common dependencies in Turkish discourse}

\author[1]{Deniz Zeyrek}[%
orcid=0000-0001-9248-0141,
email=dezeyrek@metu.edu.tr,
url=http://users.metu.edu.tr/dezeyrek/,
]
\address[1]{Middle East Technical University,
  Graduate School of Informatics, Cognitive Science Department, Dumlupınar Boulevard, No:1, 06800, Ankara, Turkey}

\author[1]{Mustafa Erolcan Er}[%
orcid=0000-0002-3009-4517,
email=erolcan.er@metu.edu.tr, url=https://github.com/erolcan-er,
]

\begin{abstract}
  We describe Turkish Discourse Bank 1.2, the latest version of a discourse corpus annotated for explicitly or implicitly conveyed discourse relations, their constitutive units, and senses in the Penn Discourse Treebank style. We present an evaluation of the recently added tokens and examine three commonly occurring dependency patterns that hold among the constitutive units of a pair of adjacent discourse relations, namely, shared arguments, full embedding and partial containment of a discourse relation. We present three major findings: (a) implicitly conveyed relations occur more often than explicitly conveyed relations in the data; (b) it is much more common for two adjacent implicit discourse relations to share an argument than for two adjacent explicit relations to do so; (c) both full embedding and partial containment of discourse relations are pervasive in the corpus, which can be partly due to subordinator connectives whose preposed subordinate clause tends to be selected together with the matrix clause rather than being selected alone. Finally, we briefly discuss the implications of our findings for Turkish discourse parsing. 
  
\end{abstract}


\maketitle

\section{Introduction}

\label{sec:intro}

Turkish is a language of more than 80M speakers and belongs to the Turkic sub-family of the Altaic language family. It is a free word-order, agglutinating language with a complex morphology, where suffixation is a major tool of both derivation and inflection.

The existing Natural Language Processing (NLP) methods for Turkish have been developed primarily targeting its morphology and syntax, lately extending to semantics \cite{eryiugit2008dependency}, \cite{ccakici2018wide}. But there is also need for discourse processing research, i.e. NLP beyond the boundaries of the sentence, which would inform systems such as information retrieval, dialogue systems, summarization. The first annotated discourse corpus of Turkish, Turkish Discourse Bank, or TDB \cite{zeyrek2013turkish} has been developed to fill the gap in the discourse processing of Turkish and is expected to support language technology applications that need information at the discourse level. It is a manually annotated corpus of modern Turkish that follows the rules and principles of the Penn Discourse Bank (PDTB) \cite{prasad2008penn} annotating discourse relations over texts from various genres (fiction, biography, newspaper editorials, popular magazines, etc.).\footnote{The earliest version, TDB 1.0, is a $\sim 400.000$-word corpus available at \url{https://github.com/disrpt/sharedtask2019/tree/master/data/tur.pdtb.tdb}. TDB 1.1, a 40.000-word-version with fewer annotations, is available at: \url{https://github.com/disrpt/sharedtask2021/tree/main/data/tur.pdtb.tdb.}} 

While the PDTB still remains the largest resource, the creation of PDTB-style discourse corpora in languages such as Turkish, Hindi, Arabic and Chinese (see \cite{prasad2014reflections} and the references therein) has been significant for empirical purposes and for discourse processing studies on those languages. The empirical value of new resources is high because they underscore both the variability and similarity of discourse-related phenomena across languages and enable researchers to reach a better understanding of discourse structure.

 The goal of the current paper was twofold: (a) To describe the latest version of TDB, namely TDB 1.2, a 40.000-word corpus, and evaluate the recently added tokens, (b) to highlight three commonly occurring discourse dependencies found in TDB 1.2; i.e., \textit{shared arguments}, \textit{full embedding} and \textit{proper containment of a discourse relation}, and discuss the issues revolving around these dependencies from the viewpoint of a morphologically rich language.

The layout of the paper is as follows. We start with an overview of the notions that underlie TDB, describe major annotation categories, and offer an evaluation of the new discourse relation tokens (\S \ref{sec:Turkish Discourse Bank 1.2}).  In \S \ref{sec:dependencies} we present the most common dependencies in the corpus and discuss the linguistic issues surrounding them, and in \S \ref{sec:conc}, we conclude the paper.

\section{Turkish Discourse Bank 1.2}
\label{sec:Turkish Discourse Bank 1.2}
\subsection{What is discourse and what are discourse relations?}
Discourse is the level of language above the sentence and can be found even within a sentence. The assumption in discourse research is that a stretch of text is not an arbitrary sequence of sentences but a structured, coherent unit that has a meaning more than the sum of its parts. Discourse structure can be discovered by examining the patterns in multi-sentence or multi-clausal texts and by finding the constitutive units of these patterns. This is essential for correctly interpreting the text \cite{webber2011discourse} and for the first step of discourse processing, i.e. discourse segmentation, known as discourse parsing. One of the key aspects of discourse structure is \textit{discourse relations}  (\textit{DRs}), which denote the semantic relatedness of two text pieces at the local level, such as contrast, additive, condition.\footnote{Although discourse relations can also express the pragmatic relatedness of discourse units (e.g. claim-evidence), they are not annotated in TDB.} 

Following the PDTB's lexicalized approach to discourse relations, it is assumed that there is lexico-syntactic evidence for the existence of discourse relations. Thus, connectives are seen as a primary source of evidence for the occurrence of a discourse relation. These are expressions such as  conjunctions and adverbs (\textit{or, although, moreover}) linking clauses that have an \textit{abstract object} interpretation (propositions or eventualities) \cite{Ash93}. They are referred to as \textit{(explicit) discourse connectives (DCs)} signalling the presence of discourse relations (see example \ref{ex:explicit} in Appendix \ref{sec:appendix-2}).  

But readers do not necessarily need discourse connectives, because they can easily infer the  relation from the adjacency of textual units, lexical relations, anaphoric links, etc. These have been known as implicit relations. Furthermore, readers can add a discourse connective to an implicitly conveyed relation to make it salient -- called ``\textit{implicit discourse connectives}'' \cite{prasad2008penn} -- and can specify the textual parts of an implicit relation (see example \ref{ex:implicit} in Appendix \ref{sec:appendix-2}). Finally, implicit relations may be realized by other means, namely, through Alternative Lexicalization (AltLex), or as Hypophora, Entity Relation, as well as No Relation (more explanation and examples are provided for each relation type in Appendix \ref{sec:appendix-2}).

\subsection{What is annotated in TDB?} 
Based on the notions described in \S \ref{sec:Turkish Discourse Bank 1.2}, three major aspects of discourse are annotated in TDB 1.2: (a) Discourse relations conveyed explicitly or implicitly as well as by other means, (b) constitutive units of discourse relations, which are known as \textit{arguments}, (c) the sense of explicitly and implicitly conveyed relations and AltLexes. There are always two textual units that constitute a relation. The textual unit syntactically hosting the discourse connective is called Argument 2, the other argument is named as Argument 1.\footnote{At least two annotators, who were graduate students at Middle East Technical University, Cognitive Science Department, were involved in each annotation cycle. The annotations were regularly checked and adjudicated by the research team.}

Although all languages have elements that function as discourse connectives, the syntactic class to which they belong may differ.  For example, Turkish not only has lexical connectives (\textit{and, but, so}) as most languages do but also converbial 
and postpositional connectives, grouped as subordinators. These connectives relate a non-finite subordinate adverbial clause to the matrix clause. In converbial structures, the marker of the relation is merely a suffix, called suffixal connectives here, which generally correspond to subordinating conjunctions in English. In postpositional structures, the marker of the relation has two parts, a postposition and a nominalization suffix on the subordinate verb. Converbial suffixes and postpositions are annotated as explicit discourse connectives in TDB.   

 Importantly, the neutral order of  arguments to subordinators is Argument2-Argument1 (i.e. the argument that hosts the connective, which is the second argument, is normally \textbf{preposed}). Both subordinator types are typically translated to English with a \textbf{postposed} subordinate clause). Example \ref{ex:suffixal} presents a suffixal connective, -\textit{ince} `when' \ref{ex:ogretmen}, while \ref{ex:ogretmen} illustrates the use of a postposition \textit{-diği gibi} `as' used as a discourse connective. Both connectives relate a preposed non-finite subordinate clause to the matrix clause. In the examples throughout the paper, the discourse connective is underlined, the inferred implicit discourse connective is both underlined and put between parentheses. Argument 1 is shown in italic fonts, Argument 2 in bold fonts. Each Turkish example is translated into English and shown between single quotation marks.

\ex. \label{ex:suffixal}
\textbf{Öğrenciler gel}-\underline{ince} \textit{aşağı indi}.\\
`\textit{He came down} \underline{when} \textbf{the students arrived}'.\\

\ex. \label{ex:ogretmen}
\textbf{Ali'nin göster}\underline{-diği gibi} \textit{resim yaptım}.\\
`\textit{I drew} \underline{as} \textbf{Ali showed}'.\\

In the rest of the paper, the patterns that involve subordinator connectives will be in focus as their syntactic behaviour is peculiar to Turkish and their analysis could highlight the differences between Turkish and other languages annotated in the same style.

\subsection{Evaluation and the finalization of TDB 1.2}
\label{sec:evaluation}

TDB 1.2 currently has a total of 3870 relations, surpassing TDB 1.1 by 2014 relations (see Appendix \ref{sec:appendix-1} for the tokens recently added to the corpus). Since earlier versions of TDB 1.1 have already been evaluated, it appeared meaningful to evaluate the recently added tokens. A group of three expert annotators worked on a randomly chosen $\sim$ 42\% of the new relations (849 tokens in total) annotated since \cite{zeyrek2017tdb}. They were told to accept the annotations, revise them where needed, or reject them, suggesting a new relation token where possible. All decisions were made unanimously by them independently of the annotators who created and adjudicated the recent tokens. In calculating inter-annotator agreement (IAA) statistics, we considered the already adjudicated tokens as created by Annotator1, and the unanimously revised tokens as created by Annotator2. Thus IAA was measured between two annotators.
We measured various types of IAA as described below and obtained a high degree of agreement in each case. 
\begin{itemize}
    \item \textit{Agreement on the DRs' type of realization:} This
is defined as the number of common discourse relations (pairs of clauses specified as a discourse relation by both annotators) over
the number of unique relations, where all relations have the same type of realization \cite{zeyrek2017tdb,forbes2016extracting}. We used the exact match criterion \cite{miltsakaki2004penn} and present the results of this analysis in Table \ref{table: type agreement} in Appendix \ref{sec:appendix-3}.

    \item \textit{Agreement on senses:} The PDTB introduces a hierarchically organized semantic categorization used to tag the sense(s) of Explicit and Implicit relations and AltLexes. The sense hierarchy has four Level-1 senses (Expansion, Contingency, Comparison, Temporal), which are further refined by Level-2 senses. A third level specifies the semantic contribution of each argument \cite{prasad2008penn}. Thus, a temporal relation anchored by \textit{then} would be annotated as Temporal:Asynchronous:Precedence, while a temporal relation expressed by \textit{after} would be annotated as Temporal:Asynchronous:Succession. Following \cite{forbes2016extracting}, we calculated sense agreement on all three sense levels of the PDTB 3.0 sense hierarchy among common discourse relations using the exact match criterion. The results are listed in Table \ref{table:sense agreement} in Appendix \ref{sec:appendix-3}.

\item \textit{Agreement on argument spans:}
TDB 1.2 asks the annotators to observe the PDTB's minimality principle, which states that the extent of the arguments to a discourse connective should be as minimal as possible as needed by the sense of the relation. The annotators are not encouraged to select distant arguments to a discourse connective but they should leave out certain expressions specified in the annotation manual (e.g. attribution phrases such as \textit{he said} should be excluded).  

To evaluate the stability of the argument span annotations, we measured IAA using Cohen's Kappa \cite{cohen1960coefficient}.

The first step involves determining the boundaries of arguments, both Argument1 and Argument2. This is known as unitization of the data (\cite{artstein2008inter, yalccinkaya2010inter}). In earlier work on TDB 1.0, the data was unitized with respect to words \cite{zeyrek2013turkish}. In the current work, we unitized the data with respect to \textit{characters} by encoding each of them as 1 or 0 (selected/excluded); that is, we recorded the number of judgements a character receives for each category and calculated agreement over the data unitized in this manner. This encoding method has been considered more advantageous than the word-based encoding as it suits the agglutinating nature of Turkish better, enabling for example, the calculation of the agreement on argument spans to suffixal connectives. The agreement on each argument was measured separately. The results are given in Table \ref{table:Cohen's K on arg span} in Appendix \ref{sec:appendix-3}.

\end{itemize}

All disagreements were resolved by the research team and the remaining discourse relation tokens checked and updated where needed. The results were recorded in the data and TDB 1.2 was created.
Recently added tokens yielded a corpus with the annotation categories distributed as shown in Table \ref{table:general dist.} in Appendix \ref{sec:appendix-3}. The table reveals that the majority of the relations are implicit amounting to 62.09\% of the total number of annotated tokens as opposed to explicit relations that constitute 37.91\% of the data.

\section{Common dependencies in TDB 1.2}
\label{sec:dependencies}
TDB annotation style reflects the incremental interpretation of texts by humans. The annotators are asked to read the text sentence by sentence and annotate different realizations of discourse relations as they appear in the text, also tagging the constituents of discourse relations along with the relations' senses. Although they are not required to annotate any dependencies among discourse units, by examining the annotation files produced by this annotation style, certain dependencies can be detected, which in turn would inform us about discourse structure, ultimately supporting discourse parsers and other language technology applications. Discourse-level dependencies have been examined in PDTB 2.0 for English \cite{lee2006complexity}, over TDB 1.0 for Turkish \cite{aktacs2010discourse, demirsahin2013applicative}, and recently for Czech \cite{polakova2021discourse}. In this paper, we continue this line of research started by Lee et al. \cite{lee2006complexity}. Examining TDB 1.2 with a Python script, we investigate the dependencies among three discourse units belonging to two consecutive discourse relations related by explicit or implicit discourse connectives (other discourse relations are out of scope of our analysis). 

The object of our investigation can be represented as: $DU_1$ - DC1 - $DU_2$ - DC2 - $DU_3$. 
That is, we deal with the dependencies among three linearly ordered discourse units (DUs), where DU means any text span selected as an argument by one or both of the discourse connectives. 
The major dependency types that we find are listed in Table \ref{table:SA/PCR} in Appendix \ref{sec:appendix-3} together with the number of times each type occurs in the data.

\subsection{Shared arguments}
Shared arguments refer to multiple parenthood, a kind of dependency where $DU_2$ is shared by the right side and the left side discourse connectives without any part of the argument span being excluded (in the examples, the shared argument is shown in a double-lined frame box to distinguish it from other DUs, which are placed in a frame box). Table \ref{table:SA/PCR} shows that 632 tokens (72.48\% of the total number of shared arguments) are an argument to an \textit{implicit} $DC_1$ shared by an \textit{implicit} $DC_2$ (the Implicit-Implicit pattern in Table \ref{table:SA/PCR}). Given the high number of implicit discourse relations in the corpus, the common occurrence of shared arguments in the Implicit-Implicit pattern is not unexpected. Also, recall that TDB is a multi-genre corpus including works of fiction, where few discourse connectives tend to occur. So, the inclusion of fiction in our corpus could be one of the reasons why implicit relations occur more frequently than explicit ones, eventually leading to the frequent occurrence of arguments shared by implicit relations.  

Example \ref{ex:shared-arg} illustrates an Implicit-Implicit dependency structure, where $ DU_{2} $ is shared by two implicit relations and the shared argument is syntactically a finite clause just like other DUs in the example. Each DU of this example is a main clause expressing an independent eventuality that can take the discourse forward. This appears to be a valid reason to make them available for reselection.

\ex.
\label{ex:shared-arg} 
\small \fbox{Bu ben değildim}, \underline{(çünkü)} \fbox{\fbox{ben yere bakmazdım},} \underline{(bilakis)} \fbox{gözüne gözüne bakardım insanların}.
\small \fbox{This was not me} \underline{(because)} \fbox{\fbox{I would not look down}}, \underline{(rather)} \fbox{I would look into people's eyes}.

Given the saliency of main clauses in discourse \cite{mann1988rhetorical}, their reselection is no surprise, but are subordinate clauses shared? As already mentioned, in Turkish, postpositional and suffixal connectives anchor non-finite (preposed) subordinate clauses. Are such clauses shared or not? We found that such subordinate clauses can be shared, though very rarely. For example, we found only 6 instances where the subordinate clause of a postposition is shared. Sentence \ref{ex:shared-arg-post} presents a causal postpositional connective \textit{için} (DC2), and its subordinate clause (\textbf{görüşmeyi kabul ettiği} `accepting to meet us') reselected without its matrix clause.  Although a detailed analysis is needed to reveal the conditions under which a preposed subordinate clause ($ DU_{2} $) is shared, it appears that in \ref{ex:shared-arg-post}, annotators have interpreted the eventuality described in $ DU_{2} $ as semantically independent possibly co-occurring with the event described in the matrix clause (DU3). This could have triggered the subordinate clause to be reselected.

\ex.
\label{ex:shared-arg-post} 
\small \fbox{Bizi aray-} \underline{-arak} \fbox{\fbox{görüşmeyi kabul ettiği}} \underline{için} \fbox{çok teşekkür ediyoruz}.\\
`\small (\underline{By})\fbox{Calling us} \fbox{\fbox{he accepted to meet with us}}, \underline{it' for this reason that} \fbox{we are thankful to him}.'\\

To summarize, our analysis shows that while it is common for two adjacent implicit discourse relations to share an argument, it is much less common for two adjacent explicit relations to share an argument, and subordinate clauses of subordinators are shared on rare occasions.

\subsection{Full embedding}

Full embedding refers to cases where a discourse relation is \textit{totally} realized as the argument to the connective. It is similar to embedding in syntax and expected to occur in TDB 1.2, too.

Indeed, it is common in the corpus, as Table \ref{table:SA/PCR} (Appendix \ref{sec:appendix-3}) reveals. 

Most of the fully embedded discourse relations appear in patterns where $DC_2$ is an explicit discourse connective, either lexical of suffixal. The Implicit-Explicit pattern, for example, occurs in 59.77\% of all fully embedded instances in Table \ref{table:SA/PCR}. This is where the second argument to an \textit{implicit} $DC_1$ is a fully embedded relation anchored by an \textit{explicit} $DC_2$. 

Example \ref{ex:embedded-2} is chosen from the Explicit-Explicit pattern. It presents a suffixal discourse connective -ip `after' and its binary arguments being fully embedded as an argument to a suffixal connective on the left side, -arak `once'. In other words, the subordinate clause of -ip (\textbf{anneannesinin yanına gel}- `move to her grandmother's) is selected together with the matrix clause, as the translation also shows.

\ex.\label{ex:embedded-2} 
\small\fbox{Hukuk Fakültesini yarım bırak}-\underline{arak} \fbox{\fbox{\textbf{anneannesinin yanına gel}}-\underline{ip} \fbox{\textit{Ankara'ya yerleşmesinin}}} nedeni ...\\
`\small the reason why \fbox{ \underline{after} \fbox{\textbf{moving to her grandmother's}} \fbox{\textit{she settled in Ankara}}}\\
 \underline{once} \fbox{she quitted the Law School}' ... \\

Different from example \ref{ex:shared-arg-post}, this subordinate clause is not selected alone and a shared argument structure does not arise. The selection of the entire discourse relation seems due to a semantic reason: rather than being an independent eventuality, the event in the subordinate clause is in a sense dependent on the event described in the matrix clause: it brings about the event in the matrix clause. The preposed position of the subordinate clause and possibly its non-finiteness coupled with its semantics appears to block its selection alone as an argument. Although our annotation guidelines do not have rules regarding such subtle issues, the annotators opted to select most of the preposed non-finite subordinate clauses together with their matrix clauses (i.e. the entire discourse relation) as an argument, leading to fully embedded clauses or properly contained discourse relations, which is the next topic below.   

\subsection{Properly contained discourse relations}
Properly contained discourse relations are a subtype of fully embedded ones except that some material is left out (shown with three dots in Ex. \ref{ex:properly-contained}) (the examination of the excluded part is left for further research). Similar to fully embedded relations, properly contained relations tend to occur in the patterns where $DC_2$ is an explicit discourse connective. For example, the Implicit-Explicit pattern comprises 55.25\% of all properly contained relations.

\ex.
\label{ex:properly-contained}
\small \fbox{çarşaflarla geceden giderek terasa saklandı} \underline{(sonra)}
   \fbox{... \fbox{\textbf{çarşafları giy}}-\underline{erek}\fbox{\textit{terastan indi}}}.\\
%
\small \fbox{he hid at the terrace with the hijab} \underline{(then)}
\fbox{... \underline{after} \fbox{{\textbf{wearing the hijab}}} \fbox{\textit{he came down}}}.

In Ex. \ref{ex:properly-contained}, chosen from the Implicit-Explicit pattern, the preposed subordinate clause ($ DU_2$) and its matrix clause ($ DU_3$) are selected entirely as the second argument to $DC_1$ rather than the subordinate clause being selected alone, which would have resulted in a shared argument structure. Once again, this seems to be due to the position of the subordinate clause as well as its semantics: the event described by the preposed subordinate clause \textbf{çarşafları giy-} `wear the hijab' engenders the main event \textit{terastan indi} `he came down'; the man wears the hijab and only then, he comes down from where he is hiding (otherwise, he would be noticed by the women, as the narrative describes). These events are not inferred as independently (co-)occurring, which seems a good reason why we find a properly contained dependency structure.

In short, preposed (non-finite) subordinate clauses in Turkish seem to trigger full embedding or proper containment structures, which could be explained not only by the position and non-finiteness of the subordinate clauses but also by their semantics in relation to the matrix clauses.

\section{Summary and conclusion}
\label{sec:conc}
We introduced TDB 1.2, a corpus that annotates different realizations of discourse relations, their arguments and senses in the PDTB style, and found that the corpus contains more implicit relations than explicit ones. Then, we zoomed in three types of dependency, which revealed an asymmetry between the occurrence patterns of shared arguments on one hand and fully embedded and properly contained discourse relations on the other. Our analyses showed that arguments are shared frequently by two adjacent implicit discourse relations, but much less so by two adjacent explicit discourse relations. Instead, discourse relations conveyed by explicit connectives such as suffixal ones or postpositions tend to be selected totally as an argument to another discourse relation, mostly an implicit one.

Our findings have implications both for discourse parsing and the theoretical understanding of Turkish paving the way for comparisons with other languages towards a better understanding of discourse. While there is room for more research on both sides, the findings minimally show that the implicit discourse relation recognition task can be improved by considering shared arguments, which demonstrate, among others, that three adjacent implicit discourse relations is a highly likely sequence in Turkish discourse. Also, automatic argument span detection can be improved by considering the availability of an entire discourse relation anchored by postpositions or suffixal connectives as an argument, as fully embedded and properly contained dependency patterns reveal.

What we have not examined in this paper is whether there are other factors involved in the formation of the dependency structures described, e.g. the sense of $DC_1$ and/or $DC_2$. The investigation of such factors is left for further research.

\begin{acknowledgments}
We acknowledge the partial support of Middle East Technical University (BAP-07-04-2017-001) and thank Salih Fırat Canpolat, Deniz Dilek Bilgiç, Ozan Deniz, Ali Can Serhan Yılmaz, Zeynep Başer, Özgür Şen Bartan, Aytaç Çeltek and Murathan Kurfalı for their assistance at various stages of the development of TDB 1.2. Any remaining errors are our own. 
\end{acknowledgments}

\bibliography{sample-ceur}

\appendix

\section{Appendix: Major annotation categories and examples in TDB 1.2}
\label{sec:appendix-2}

TDB 1.2 annotates implicitly and explicitly conveyed discourse relations that hold between adjacent verb phrases, clauses, and sentences. This section illustrates major annotation categories together with examples. 

\textbf{Explicit relations} - An explicit discourse relation holds when the relation is encoded through an overt discourse connective.  

\ex. \label{ex:explicit}
\small \textbf{Ali uzun boylu} \underline{ama} \textit{kız kardeşi kısa boyludur}.\\
`\small \textbf{Ali is tall}, \underline{but} \textit{his sister is short}.' 

\textbf{Implicit relations} - In cases where an overt discourse connective is absent, an implicit discourse relation is inferred and shown by inserting a discourse connective in the relation. 
    
\ex. \label{ex:implicit}
\small \textit{Yol kaygandı,} (\underline{Imp=o yüzden}) \textbf{Ali arabayı dikkatli kullandı}.\\
`\small \textit{The road was slippery,} (\underline{Imp=due to that}) \textbf{Ali was driving carefully.'} 
    
\textbf{Alternative Lexicalization (AltLex}) - When a discourse relation is alternatively lexicalized through linguistic expressions such as \textit{despite this, because of this, the reason is}, the relation is called and AltLex.
    
\ex. \label{ex:altlex}
\small \textit{Ali Latince öğrendi.} \underline{Bundan sonra} \textbf{Fransızca kitap okumak çok kolay oldu}. \\
  `\small \textit{\textit{Ali learnt Latin}}. \underline{After that}, \textbf{\textbf{reading books in French has been so easy}}.'
  
\textbf{Entity Relation (EntRel)} - This is where the text spans express a relation with an entity.
    
    \ex. \label{ex:EntRel}
    \small \textit{Dr. Ahmet bey yeni bir hastahanede işe başladı}. \textbf{Rahmetli Dr. Ali bey'in yerini aldı.}\\
   `\small  \textit{Dr. Ahmet Beg has started to work in a new hospital}. \textbf{He succeeds the late Dr. Ali Beg.}'
    
\textbf{Hypophora} - These are questions and meaningful answers given to the questions. 

\ex.
\small \textit{Fıkra hoşuna gitti mi}? \textbf{Evet bayıldım}. \\
`\small \textit{Did you like the joke}? \textbf{Yes I loved it}.'  
    
\textbf{No Relation (NoRel)} - A NoRel involves cases where no relation can be inferred between adjacent text spans. 
    
    \ex. \label{ex:NoRel}
   `\small  \textit{Okul yakında tatile girecek}. \textbf{Öğretmenler okula gönderilmeyen öğrencilerle uğraşamaz}.'\\
    \small \textit{Children will have a break soon}. \textbf{Teachers can't deal with students not sent to school.}

Explicit and Implicit relations and AltLexes are annotated both within and across sentences, while Hypophora tokens, EntRels, and NoRels are annotated only between adjacent sentences. 

\section{Appendix: Tokens recently added to TDB 1.2}
\label{sec:appendix-1}
The most recent additions to the corpus involve implicit verb phrase conjunctions (Ex. \ref{ex:imp VP conj}) and multiple relations (examples \ref{ex:multiple-rels} - \ref{ex:multiple-altlex}).

\ex.  \label{ex:imp VP conj}
\small \textit{Çabuk değişen} (\underline{Imp=ve}) \textbf{yaşlanan} bir nüfusumuz var.\\
`\small We have a population that \textit{rapidly changes} (\underline{Imp=and}) \textbf{ages}.'

Multiple relations comprise:
\begin{itemize}

\item the implicit senses of explicitly conveyed verb phrase conjunctions (only the senses of relations marked by the conjunction \textit{ve} `and' were considered) (Ex. \ref{ex:multiple-rels}).

\item multiple relations between the same argument spans conveyed by co-occurring explicit connectives, such as \textit{ve böylece} `and hence' (Ex. \ref{ex:multiple-exp}).

\item multiple relations between the same argument spans conveyed by an explicit connective and an AltLex, such as \textit{ve buna rağmen} `and despite this' (Ex. \ref{ex:multiple-altlex}).\footnote{PDTB 3.0 annotates multiple senses for explicit or implicit relations if annotators infer more than one sense as holding between a pair of spans. In TDB 1.2, multiple senses were not annotated systematically.} 
\end{itemize}

\ex. \label{ex:multiple-rels}
\small \textit{Okulu bıraktı} \underline{ve} (\underline{Imp=sonra}) \textbf{evlendi}.\\
`\small She \textit{left school} \underline{and} (\underline{Imp=then}) \textbf{got married}.'

\ex. \label{ex:multiple-exp}
\small Ayşe sevdiğiyle evlendi \underline{ve} \underline{böylece} dünyanın en mutlu kızı oldu.\\
`\small Ayşe married her beloved one \underline{and} \underline{so} \textbf{she became the happiest women in the world.'} 

\ex. \label{ex:multiple-altlex}
\small \textit{Ali okuldan nefret etti} \underline{ve} \underline{buna rağmen} \textbf{liseden mezun olmayı başardı.}\\
`\small \textit{Ali hated school} \underline{and} \underline{despite this} \textbf{he managed to finish high school}.'

Multiple relations were annotated separately on each token as in the PDTB, then linked with the same index value in their link fields.

\section{Appendix: Summarization tables}
\label{sec:appendix-3}

\begin{table}[h!]
\caption{IAA results for agreement on DRs' type of realization}
\begin{tabular}{ll}
\hline
\textbf{Realization Type} & \textbf{Agreement} \\ \hline
Implicit               & 0.97             \\
Explicit               & 0.99             \\
AltLex                 & 0.98             \\
EntRel                 & 0.95             \\
Hypophora              & 1.00                 \\
NoRel                  & 0.97             \\ \hline
\end{tabular}
\label{table: type agreement}
\end{table}

\begin{table}[h]
\caption{IAA results for sense agreement}
\begin{tabular}{llll}
\hline
\textbf{Sense} &
  \textbf{\begin{tabular}[c]{@{}l@{}}Explicit \\ (\%)\end{tabular}} &
  \textbf{\begin{tabular}[c]{@{}l@{}}Implicit \\ (\%)\end{tabular}} &
  \textbf{\begin{tabular}[c]{@{}l@{}}AltLex\\  (\%)\end{tabular}} \\ \hline
\textbf{Level-1} & 99.02 & 99.75 & 100     \\ 
\textbf{Level-2} & 98.66 & 99.75 & 100     \\ 
\textbf{Level-3} & 80.11 & 79.44 & 79.83 \\ \hline
\end{tabular}
\label{table:sense agreement}
\end{table}

\begin{table}[h!]
\caption{IAA results for argument span selection}
\begin{tabular}{ll}
\hline
\textbf{Arg Type} & \textbf{Cohen's $\kappa$} \\ \hline
Argument1              & 0.90                 \\
Argument2              & 0.85                 \\ 
\hline

\end{tabular}
\label{table:Cohen's K on arg span}
\end{table}

\begin{table}[h!]
\caption{Number of different realizations of discourse relations and their Level-1 sense tags in TDB 1.2}
\begin{tabular}{lllllll}
\hline
               & \textbf{Expansion} & \textbf{Temporal} & \textbf{Comparison} & \textbf{Contingency} & \textbf{DRs with no sense tag} & \textbf{Total} \\ \hline
\textbf{Implicit}  & 1090 & 158 & 162 & 333 & 0   & \textbf{1743} \\
\textbf{Explicit}  & 540  & 400 & 259 & 268 & 0   & \textbf{1467} \\
\textbf{AltLex}    & 33   & 32  & 14  & 67  & 0   & \textbf{146}  \\
\textbf{EntRel}    & 0    & 0   & 0   & 0   & 233 & \textbf{233}  \\
\textbf{Hypophora} & 0    & 0   & 0   & 0   & 78  & \textbf{78}   \\
\textbf{NoRel}     & 0    & 0   & 0   & 0   & 203 & \textbf{203}  \\
\textbf{Total} & \textbf{1663}      & \textbf{590}      & \textbf{435}        & \textbf{668}         & \textbf{514}      & \textbf{3870}  \\ \hline
\end{tabular}
\label{table:general dist.}
\end{table}

\begin{table}[h!]
\caption{Number of common dependencies in TDB 1.2}
\resizebox{\textwidth}{!}{%
\begin{tabular}{lllllll}
\hline
DC1 & Explicit & Explicit & Implicit & \multirow{2}{*}{\textbf{Sub Total}} & Implicit & \multicolumn{1}{c}{\multirow{2}{*}{\textbf{Total}}} \\
DC2                    & Explicit & Implicit & Explicit &      & Implicit & \multicolumn{1}{c}{} \\ \hline
Shared Arguments       & 41       & 105      & 96       & 240  & 632      & 872                  \\
Fully embedded DRs     & 117      & 85       & 471      & 673  & 115      & 788                  \\
Properly Contained DRs & 145      & 82       & 521      & 748  & 195      & 943                  \\ \hline
\textbf{Total}         & 303      & 272      & 1088     & 1663 & 942      & 2605                 \\ \hline
\end{tabular}%
}
\label{table:SA/PCR}
\end{table}

\end{document}